\documentclass[letterpaper, 10 pt, journal,twoside]{ieeeconf}  %
\let\labelindent\relax

\IEEEoverridecommandlockouts                              %

\usepackage{graphicx}
\usepackage{enumitem}
\usepackage{booktabs}
\usepackage{algorithm}
\usepackage[noend]{algpseudocode}
\usepackage{multirow}
\usepackage{svg}
\usepackage{subcaption}
\usepackage{cite}
\usepackage{hyperref}
\renewcommand{\baselinestretch}{0.94}
\usepackage{xcolor}

\usepackage{flushend} %

\title{InteLiPlan: An Interactive Lightweight LLM-Based Planner\\for Domestic Robot Autonomy}

\author{Kim Tien Ly$^{1}$, Kai Lu$^{2}$, Ioannis Havoutis$^{2}$%
\thanks{Manuscript received: September 2, 2025; Revised: November 30, 2025; Accepted: January 8, 2026.}
\thanks{This paper was recommended for publication by Editor Jens Kober upon evaluation of the Associate Editor and Reviewers' comments. This work was supported by the EPSRC project EP/Z531212/1.}
\thanks{
$^{1}$: K. T. Ly is with Middlesex University.  Email: \tt \small{k.ly@mdx.ac.uk}.}
\thanks{$^{2}$: K. Lu and I. Havoutis are with the Oxford Robotics Institute, University of Oxford. Email: \tt \small{\{kailu,ioannis\}\allowbreak@robots.ox.ac.uk}.} 
\thanks{Digital Object Identifier (DOI): see top of this page.}
}

\begin{document}
\bstctlcite{IEEEexample:BSTcontrol}
\maketitle
\markboth{IEEE Robotics and Automation Letters. Preprint Version. Accepted January 2026.}{Ly \MakeLowercase{\textit{et al.}}: InteLiPlan: An Interactive Lightweight LLM-Based Planner for Domestic Robot Autonomy}
\pagestyle{headings}

\begin{abstract}
   We introduce an interactive LLM-based framework designed to enhance the autonomy and robustness of domestic robots, targeting embodied intelligence. 
   Our approach reduces reliance on large-scale data and incorporates a robot-agnostic pipeline that embodies an LLM. Our framework, \textit{InteLiPlan}, ensures that the LLM's decision-making capabilities are effectively aligned with robotic functions, enhancing operational robustness and adaptability, while our human-in-the-loop mechanism allows for real-time human intervention when user instruction is required. We evaluate our method in both simulation and on the real robot platforms, including a Toyota Human Support Robot and an ANYmal D robot with a Unitree Z1 arm.
   Our method achieves a 95\% success rate in the `fetch me' task completion with failure recovery, highlighting its capability in both failure reasoning and task planning. \textit{InteLiPlan} achieves comparable performance to state-of-the-art LLM-based robotics planners, while using only real-time onboard computing. 
   Project website:  \href{https://kimtienly.github.io/InteLiPlan}{\textcolor{blue}{https://kimtienly.github.io/InteLiPlan}}.
\end{abstract}

\IEEEpeerreviewmaketitle

\section{Introduction}

\PARstart{I}{n} recent years, incorporating large language models (LLMs) and vision-language models (VLMs) into robotic systems as high-level planners has made remarkable progress in the Embodied AI field, with applications in domestic autonomy and robotic assistance. In the hierarchical multimodal frameworks, LLMs or VLMs serve as central components for scene understanding and reasoning, as well as generating action steps. 
Exploiting the reasoning abilities of these advanced models, LLM-based frameworks have shown promise in enhancing human-robot interaction and enabling intuitive decision-making in robotics.

Despite their impressive capabilities, applying LLMs in robotics presents unique challenges \cite{wang2024large}. These challenges stem primarily from the constraints of robot kinematics and the dynamic nature of the environments in which robots operate. 
Additionally, integrating these generative models like SayCan\cite{ahn2022can} within the existing robotics pipelines is non-trivial, as these are usually tightly coupled systems of perception, planning, and actuation. Similarly, large vision-language-action (VLA) models, such as RT-2 \cite{brohan2023rt} and $\pi_{0.5}$\cite{intelligence2025pi05visionlanguageactionmodelopenworld}, require massive datasets or offboard resources.
Having such systems running in real-time robotic deployment also poses a significant challenge due to the computational overhead.
Moreover, robotic motion safety and robustness are critical concerns in human-robot environments.
Therefore, developing a general-purpose autonomous robotic system for domestic scenarios remains a challenging and open problem.

\begin{figure}
    \centering
    \includegraphics[width=0.9\linewidth]{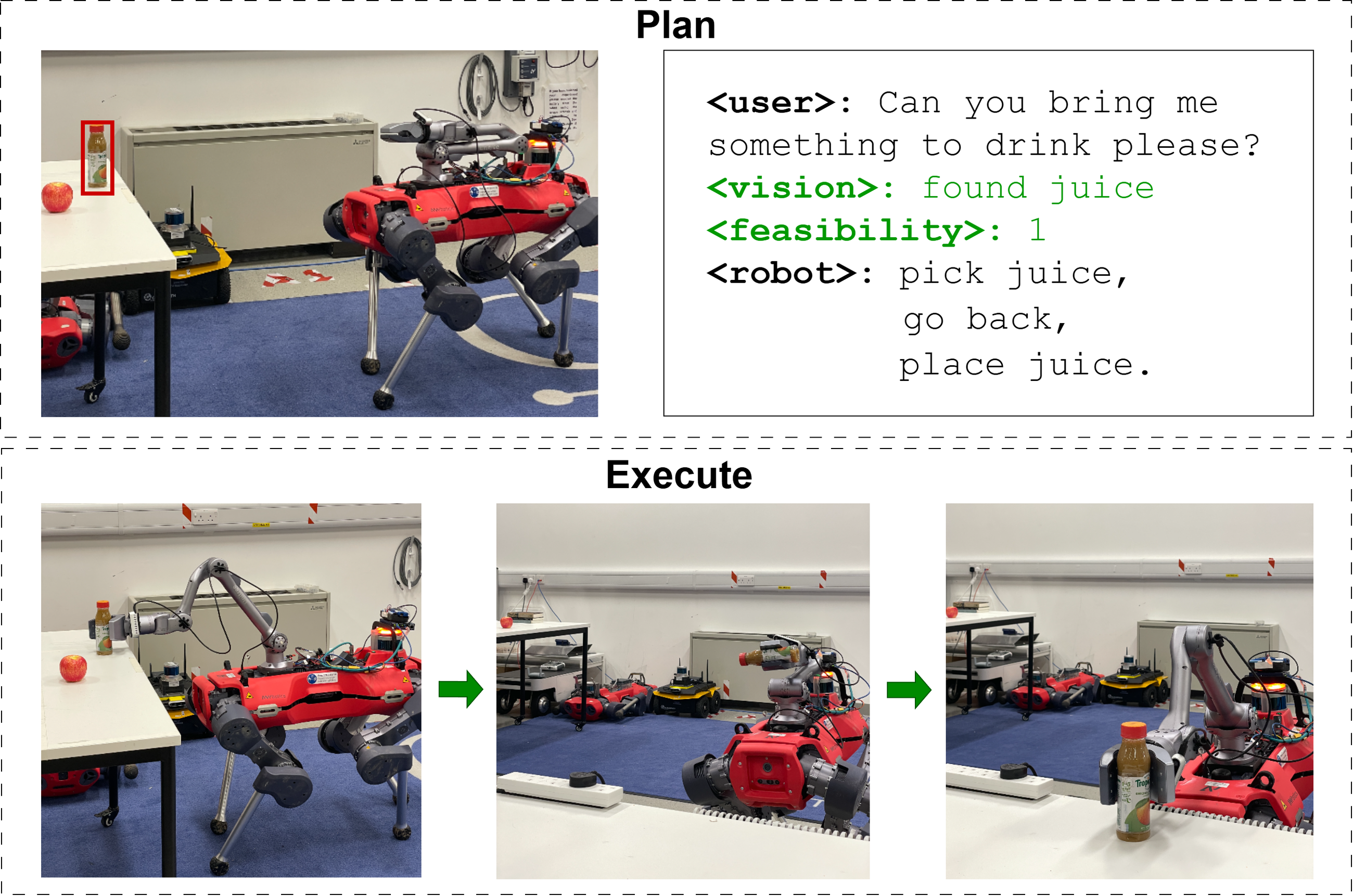}
    \caption{Step-by-step execution of \textit{InteLiPlan} result on the physical Anymal D-Unitree Z1. In our system, the robot receives either requests or guidance from humans (\textless user\textgreater). Our lightweight onboard planner will generate the robot actions (\textless robot\textgreater) considering the success of targeted object detection (\textless vision\textgreater), whole-body feasibility score (\textless feasibility\textgreater) and execute them sequentially.}
    \label{fig:physicalhsr}
\vspace{-5mm}
\end{figure}

In this work, we propose a lightweight LLM-based framework designed for domestic robots, aiming to address the aforementioned challenges. Our approach reduces reliance on large-scale data and develops a robot-agnostic pipeline. 
Furthermore, we detail strategies to mitigate kinematic constraints through real-time feasibility checks, ensuring that decision-making processes facilitated by the LLM align effectively with the physical capabilities of robots.

An integral part of our framework is the human-in-the-loop (HITL) mechanism, which allows the system to handle system failures (e.g., partially observable environment, vision failures due to lighting, etc.) or task confusions (e.g., ambiguous references or multiple candidate objects) through human intervention. 
This feature is critical for maintaining autonomy, especially in complex environments where decisions by the LLM may fail to reason what the user requested from an embodied perspective. 
The system allows human operators to input corrective actions directly, facilitating on-the-fly adaptation to unanticipated situations. 
This real-time interaction not only enhances the robustness and reliability of robotic operations but also enables adjusting the LLM's responses in accordance with human guidance.

By tackling these critical issues, our framework not only enhances the practical applicability of LLMs in robotics but also paves the way for more intuitive and responsive robotic systems capable of complex interactions and tasks in varied real-world scenarios. In summary, our work investigates the following questions:
\begin{itemize}[noitemsep]
    \item How can we design a lightweight LLM-based planner that effectively integrates user commands, visual perception, and action feasibility with onboard computing resources?
    \item How can robots use LLMs to translate human-language inputs to reliable and executable action directly? 
    \item How can uncertainties in robotic planning be solved through human intervention?
    \item How can reachability-aware LLM-based robotic planners be applicable across embodiments without additional retraining?
\end{itemize}

\section{Related work}

\subsection{LLMs for Robotic Systems}

The application of LLM models in robotics, driven by recent advancements in natural language processing, has become a focal point of current research .
Silver et al. \cite{silver2022pddl}  and Plansformer \cite{pallagani2022plansformer} use pretrained LLMs for PDDL-style planning, but still require handcrafted domain specifications, which limits extensibility for end users. In contrast, we aim to bridge human language and formal task descriptions without relying on explicit planning domains.

Liang et al. \cite{liang2023code} use an LLM for generating code policies that can be executed on the robot. However, they do not account for the robot's kinematics or geometry in the planning process, which is critical in real-world applications.
Conversely, SayCan \cite{ahn2022can} and Text2Motion \cite{lin2023text2motion} are notable for considering geometric feasibility when planning the action sequence. 
PIGINET \cite{yang2022sequence} instead introduces a transformer-based plan feasibility predictor to be integrated in a TAMP planner. The method fuses six cameras, making it harder to deploy in domestic settings. Our research diverges from these models by a multimodal structure that processes user input, visual and geometric information to output a sequence of actions that directly interface with the motion APIs.

More recently, vision-language-action models (VLAs) have emerged as another promising direction, such as RT-2\cite{brohan2023rt} and $\pi_{0.5}$\cite{intelligence2025pi05visionlanguageactionmodelopenworld}, but they require large robot-specific datasets, which can make deployment more challenging. Our approach instead develops a lightweight, robot-agnostic solution using textualized multimodal data.

\subsection{Human-Robot Interaction for Autonomy}

Recent research indicates that LLMs may not fully replicate human reasoning capabilities \cite{kaddour2023challenges}. Consequently, integrating HITL is essential for maintaining the reliability and safety of robotic systems \cite{zhang2023large}.  Vemprala et al. \cite{vemprala2023chatgpt} implement ChatGPT with HITL, emphasizing the dual benefits of enhanced model training through human feedback and increased safety during operations. Ren et al. \cite{ren2023robots} focus on human-robot interaction, where the robots decide whether or not to ask for human intervention. The framework, however, remains a sole task planner that evaluates task reasoning without considering the actual robot constraints. The decision to ask the human or not depends on the planner's confidence in generating a logical solution rather than an explicit failure. Moreover, the developed algorithm also relies on a scene description from the prompt and does not have any perception input. Our approach underlines the importance of human oversight in critical decision-making problems, and facilitates a clear and intuitive interface where the human is informed during robot operation.

\subsection{Replanning for Failure Recovery}

Failure recovery has been widely addressed in the literature for robotics planners. When it comes to LLM-based solutions, researchers typically address replanning from a second-level action check when the plan is generated \cite{ahn2024vader, joublin2024copal}. Vision-language models are also implemented to describe the error from the scene \cite{wang2023describe}. 
REFLECT \cite{liu2023reflect} incorporates visual and audio sensory data to explain failures from the failed actions. This showcases the success of failure reasoning with a multimodal structure, however, it does not consider robotic constraints and remains a pure task planner.
ReplanVLM \cite{mei2024replanvlm} uses three different VLM agents for planning, replanning with internal and external errors, indicating that a robust solution is hard to obtain by prompting one single LLM planner. 
CAPE \cite{raman2024cape} instead proposes re-prompting to replan upon failure with preconditions similar to classical planning methods (e.g., PDDL, STRIPS). Considering the context of task and motion planning, this method works in the form of hierarchical layers and replans upon execution failure. In addition to recovering from execution, we designed a multimodal LLM-based planner that incorporates action feasibility during generating plans, which helps to reduce the time taken for re-querying the planner.

\subsection{Data-efficiency in LLM-based robotics planner}

LLM-based models are usually data-intensive and require substantial training datasets and computing resources to generate robust solutions.
In robotics, this demand poses a significant challenge, as collecting large, diverse, and representative datasets from physical robots is often prohibitively time-consuming, labor-intensive, and costly.
For instance, RT series \cite{brohan2022rt,brohan2023rt}, OpenVLA \cite{kim2024openvlaopensourcevisionlanguageactionmodel,kim2025finetuningvisionlanguageactionmodelsoptimizing} and $\pi_{0.5}$ \cite{intelligence2025pi05visionlanguageactionmodelopenworld} introduces an end-to-end approach to applying transformers to robotics, requiring a large amount of data collected in months to map vision and language to actions.
Similarly, Lynch et al. \cite{lynch2023interactive} proposes an interactive method for language-conditioned robot skills, compiling a dataset through 2.7k hours of collection on real robots. Prompt engineering, on the other hand, has shown promise in applying LLMs to robotic tasks in a zero-shot manner \cite{huang2022language,huang2022inner,brown2020language}. However, this is often not robust for real-world execution without considering and understanding the operating environment. Parameter-efficient fine-tuning (PEFT) \cite{ding2023parameter} techniques have become a promising approach to minimize fine-tuning costs while enabling LLMs to acquire task-specific parameters.
Our work employs this method to incorporate robotics knowledge to LLMs while introducing a widely applicable LLM-based framework that is easy to deploy on different robot platforms,
as it does not require robot-specific data.

\section{Methodology}

\begin{figure*}[ht]
\vspace{1mm}
\setlength{\abovecaptionskip}{1mm}
\centering
\begin{subfigure}[t]{0.65\textwidth} %
    \centering
    \includegraphics[trim={12.05cm 8cm 12cm 8cm},clip,width=\textwidth]{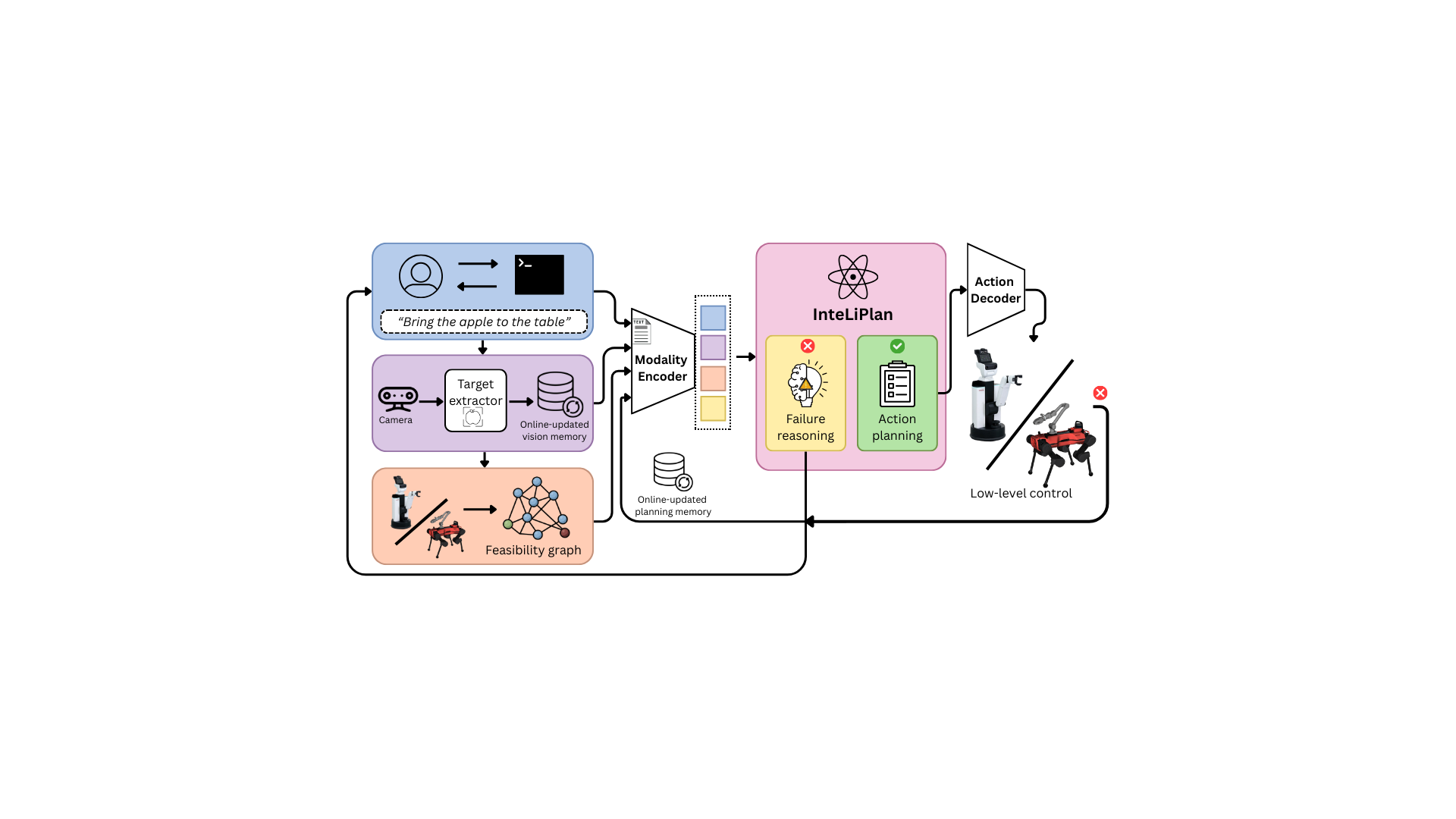}
    \label{fig:system}
\end{subfigure}
\hfill
\begin{subfigure}[t]{0.342\textwidth} %
    \centering
    \includegraphics[width=\textwidth]{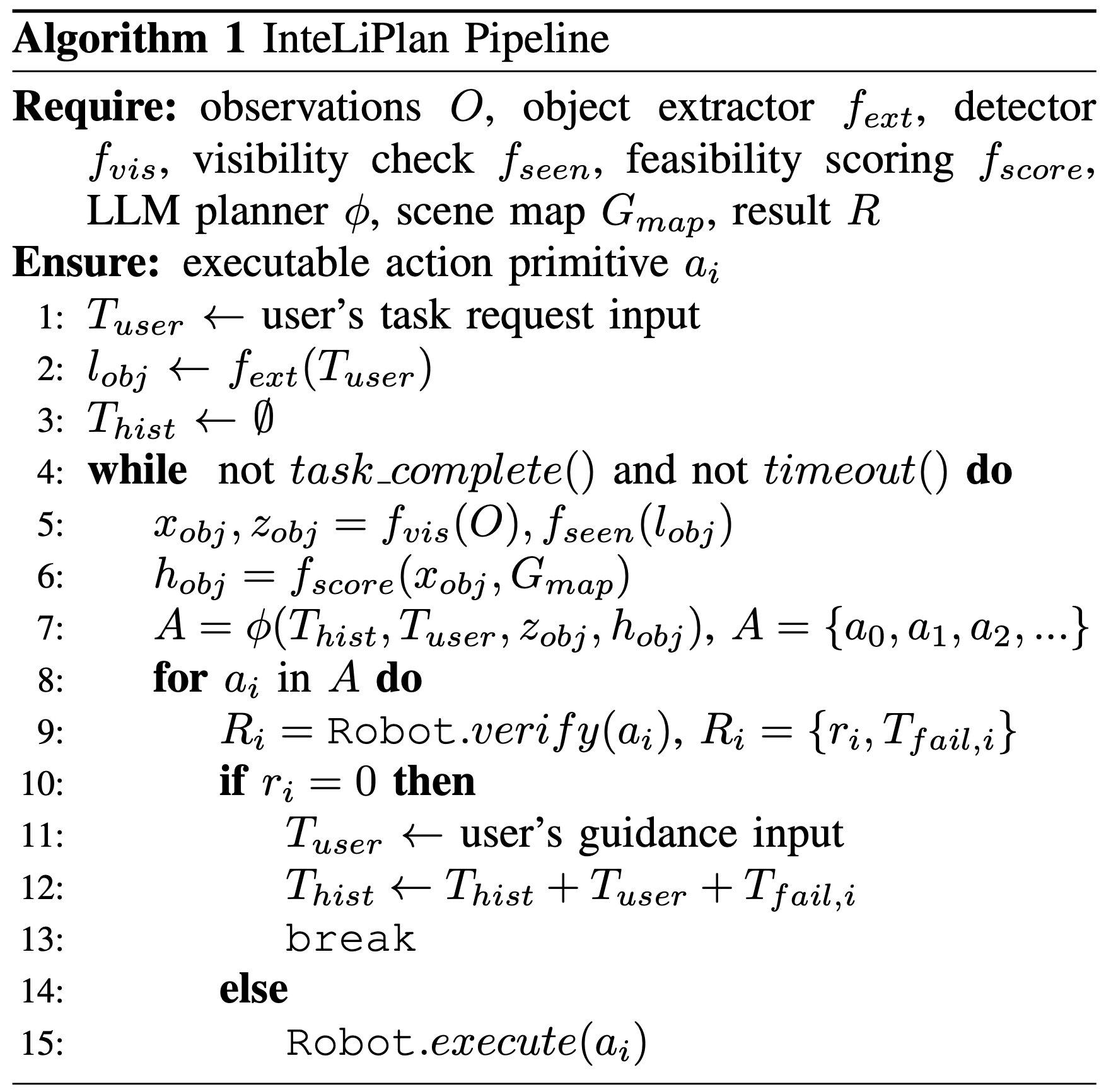}
    \label{fig:second}
\end{subfigure}

\vspace{-3mm}
\caption{System overview. Our multimodal planner integrates the user’s textual command/intervention, visual perception, and action feasibility score as inputs to a fine-tuned lightweight LLM. The LLM will then generate an action sequence for a real robot to perform mobile manipulation tasks.}
\label{fig:system}
\vspace{-5mm}
\end{figure*}
\subsection{Problem Statement}

Our work involves a mobile manipulator interacting with non-expert users to perform a human request. \textit{InteLiPlan} shares the same set of skills with the motion planner, including action primitives such as \textit{go}, \textit{pick}, \textit{place}, \textit{open}, \textit{close}, \textit{search}, and \textit{turn}. At runtime, the planner outputs an action sequence $A=(a_0, a_1, a_2, ...)$, where $a_i$ is an action from the shared skillset.
During operation, the robot receives textual inputs $T$ from a human user, which is a request at first, or guidance for recovery. The centralized planner, \textit{InteLiPlan}, takes in the user input and the internal vision and motion verification functions, in order to generate a feasible plan. If a plan is found, the sequence of actions will be executed on the robot. If a failure is detected, the robot will notify the reason and ask for human instruction. 

The robot obtains environmental information from a vision module, processing input from the onboard RGB-D camera. The visual detection
is represented as $O$. The motion verification module assesses the feasibility score of the action during planning. The framework assumes standard robot prerequisites including an environment map, skill primitives, and an object detector. 
In this setting, we define \textit{InteLiPlan} as a real-time onboard LLM-based planner, with the ability to interact with humans in natural
language and replan upon guidance. We evaluate \textit{InteLiPlan} on domestic tasks, aiming at a human-centric and interpretable framework.

\subsection{Interactive Planner with Multimodal Perception}

Our system processes multimodal inputs from the user $T_{user}$, visual observations $O$, and motion feasibility scores $f_{score}$. At the core of our system is a fine-tuned LLM that serves as the interactive planner, as illustrated in Fig. \ref{fig:system}.
Algorithm in Fig. \ref{fig:system}  explains the internal interaction between the modules.
First, user commands are passed to the vision module, which detects objects in real time and maintains an updated list of their positions. The system then checks whether the referenced items appear in the current observation.
If the object is found, the motion verification module gets the object position and checks if the target object is reachable. For example, in our implementation, we use the reachability graph designed in R-LGP\cite{ly2023r}, which is used as a feasibility check service for logic-geometric programming (LGP) framework \cite{toussaint2015logic}. This reachability graph takes in the desired end-effector position (e.g., object location $x_{obj}$) and current end-effector position and determines whether or not there is a whole-body collision-free trajectory between the two end points, regarded as $f_{score}(x_{obj},G_{map})$. In InteLiPlan, we incorporate a binary score for feasibility check, which applies, but is not limited to the reachability graph in R-LGP. Lastly, we keep track of the conversation by inserting a history $T_{hist}$ to each planner call. $T_{hist}$ is initialized empty and records the initial request with the most recent failure detected by the planner. When multiple failures occur, $T_{hist}$ will replace the failure with the immediate one, which helps to reduce memory loads and inference time.

Bringing together multimodal observations and interaction history, we formally define \textit{InteLiPlan} as:
\begin{equation}
    A =
\phi (T_{hist}, T_{user}, f_{seen}(l_{obj}), f_{score}(x_{obj},G_{map})),
\end{equation}
where $A =(a_0, a_1, a_2, ...)$ represents the output sequence of actions by LLM planning. While $T_{user}$ is the direct input from the user, $T_{hist}$ is a quoted string that tracks the previous conversation for contextual information. We define $f_{seen}(\cdot)$ as a textual feedback from the visual observation checking the list of objects $l_{obj}$ extracted from the initial request, and $f_{score}(\cdot)$  as a binary signal indicating the feasibility of the action targeted at a specific object.

Our framework features a centralized LLM-based planner to utilize the integration and coordination between submodules. We provide a plug-and-play modular design to leverage state-of-the-art models from different areas (e.g., object recognition) in a zero-shot way.

\subsection{Failure Recovery}
The ability to replan task and motion upon failure is a critical feature towards real-world applications. This can be accomplished by active exploration, or with guidance from humans. 
While prior work such as \cite{shirasaka2023self} studies self-recovery with LLM-based planners, our focus is on scenarios where human intervention is necessary, e.g., when the task is unclear, user preferences of choices are required, or the robot is limited by partial observability.
To address these cases, we incorporate a human-in-the-loop (HITL) mechanism that leverages human input for corner cases. 
By integrating human insights, \textit{InteLiPlan} can navigate complex environments more effectively and perform tasks with greater precision, thereby increasing the system's overall autonomy and operational safety.

\begin{figure*}[ht]
\vspace{2mm}
\setlength{\abovecaptionskip}{2mm}
  \centering
  \includegraphics[width=0.95\linewidth]{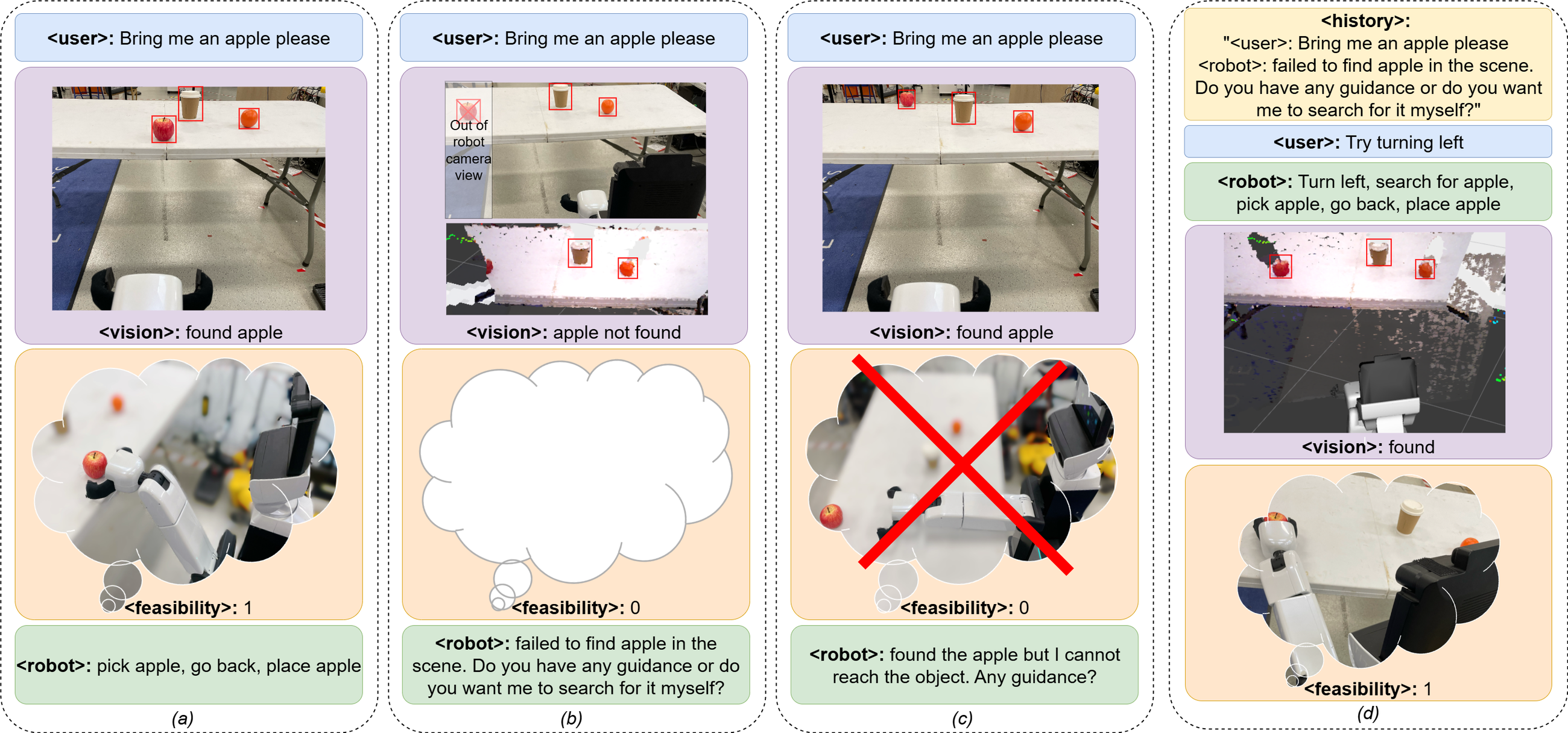}
  \caption{Examples of the multimodal LLM-based planner. (a) presents a no-failure case for the given command. (b) and (c) depict the failure reasoning ability of the model considering the inputs, for vision and feasibility failures respectively. (d) showcases the ability to recover from the failure in (b) with human instruction. In the replanning case (d), \textless history\textgreater  tracks the initial request and the latest failure found.}
  \label{fig:examples}
  \vspace{-4mm}
\end{figure*}

We consider three main categories of failures: planning failure, task confusion and execution failure, as outlined in Table \ref{tab:failedcases}. 
The multimodal structure allows the planner to reason around failures through feedback from the vision and motion verification modules, extracting the missing information to complete the task. When a failure occurs, our system can report to the human with a contextual description, interpreting the current observation and the failure that occurred.
With fine-tuning, the LLM can effectively process human inputs and replan based on human guidance. For instance, vision failures due to partial observation or adverse lighting conditions can be mitigated with human insights, leveraging the planner's capability to maintain a record of initial commands through the internal dialogue $T_{hist}$ and to replan upon system failure as necessary. This capability significantly enhances transparency, robustness, and reliability. By injecting fine-tuning data, the output is guaranteed to consider necessary observations before executing actions.

\begin{table}[h]
    \centering
    \caption{Cases that require human interaction.}
    \vspace{-1mm}

    \begin{tabular}{p{2cm}|p{2cm}|p{3.2cm}}
        \toprule
           Category & Case & Examples \\
         
         \midrule
         \multirow{2}{*}{Planning failure} & Vision failure & Failed to find object \\
         & Feasibility failure & Unable to reach object \\
         \midrule
         
         \multirow{2}{*}{Task confusion} & Ambiguous reference & Found more than 1 specified object \\
         & Ambiguous task & The user asks for "something to drink"\\
         \midrule
         
         Execution failure & Action failure & Ask for recovery \\
         \bottomrule
         
    \end{tabular}
    \label{tab:failedcases}
\vspace{-2mm}
\end{table}

\subsection{Textualized Multimodal Fine-Tuning Data for Robotics Planner}
\label{sec:data}

Recently, prompt engineering has become popular for implementing ready-to-use LLM-based planners, but poses a risk of plan reliability when executing on embodied agents. In addition, outputs from such planners often need translation into robotic control commands. Our method implements few-shot fine-tuning to inject the multimodal structure to the LLM, as well as inform the planner of the desired output format that enables direct calls to the action API. This approach also lays out the context and reduces the need for additional conversion, which uses computing resources that can prevent real-time planning.

To fine-tune the LLM model, we collected a pure-text customized dataset that includes scenarios both with and without failure recovery, where the interactions in failure cases are limited to a maximum of two rounds. The dataset is formulated as $D = \{T_{hist},T_{user},f_{seen},f_{score},A\}$. We loop through a list of objects and commands to generate the command and expected sequence of actions. Visual observation $f_{seen}$ and feasibility score $f_{score}$ will be processed by the fixed visual module and feasibility module to generate textual signals.
This text-based-only fine-tuning approach enables the model to easily adapt to different robots without further modifications or re-parameterization.  The multimodal inputs are labeled using specific flags: \textit{\textless history\textgreater} for $T_{hist}$, \textit{\textless user\textgreater} for $T_{user}$, \textit{\textless vision\textgreater} for $f_{seen}$, \textit{\textless feasibility\textgreater} for $f_{score}$, where the output label is \textit{\textless robot\textgreater} for $A$. One example of the multimodal conversation is shown in Fig. \ref{fig:physicalhsr}.

Our methodology employs Low-Rank Adaptation (LoRA) \cite{hu2022lora} to fine-tune LLM efficiently. Instead of updating all model parameters, LoRA introduces trainable low-rank matrices into key projection layers, significantly reducing computational and memory overhead. LoRA is one of the PEFT \cite{ding2023parameter} techniques that has also been proven to be superior over full fine-tuning as a parameter-efficient fine-tuning method in \cite{weyssow2023exploring}. We implement supervised fine-tuning with the collected dataset, where $\{T_{hist},T_{user},f_{seen},f_{score}\}$ is the input, and $A$ is the desired output, with cross-entropy loss. This approach enables efficient adaptation of the model to domain-specific tasks while maintaining scalability.

\section{Experimental Results}
\label{sec:result}

\subsection{Evaluation of Robustness Across Different Failures}

We first evaluate our approach using the Toyota Human Support Robot (HSR) in a `fetch me' task, where the robot needs to bring a requested item to the human. The environmental setup includes 10 seen objects and 20 unseen objects, and the planner is evaluated on 4 seen request commands, 23 unseen request commands, and replan with 9 seen guidance commands, and 18 unseen guidance commands. The unseen commands represent the variants in human-like conversation.

\begin{table}[t]
    \vspace{1mm}
    \caption{Success Rates (\%) of Task Planning.}
    \vspace{-1mm}
    \centering
    \resizebox{\linewidth}{!}{
    \begin{tabular}{c|c|c|c}
        \toprule
        Scenario & Method & w/o failure  & w/ failure \\
         \midrule
        \multirow{4}{*}{Seen cmd + Seen obj} & Deepseek-Prompt & 100 & 100\\ 
         & Deepseek-Ours & 100 & 100\\
         & Mistral-Prompt & 100 & 52 \\ 
         & Mistral-Ours & 100 & 100\\
         \midrule
        \multirow{4}{*}{Seen cmd + Unseen obj} & Deepseek-Prompt & 90 & 35\\ 
         & Deepseek-Ours & 100 & 86  \\
         & Mistral-Prompt & 90 & 29\\ 
         & Mistral-Ours & 100 & 90\\
         \midrule
         \multirow{4}{*}{Unseen cmd + Seen obj} & Deepseek-Prompt & 98.95 & 54\\ 
         & Deepseek-Ours & 100 & 93\\ 
         & Mistral-Prompt & 86.32 & 49\\ 
         & Mistral-Ours & 100 & 94\\
         \midrule
         \multirow{4}{*}{Unseen cmd + Unseen obj} & Deepseek-Prompt & 95.79 & 39\\ 
         & Deepseek-Ours & 97.89 & 93\\ 
         & Mistral-Prompt & 78.95 & 36\\
         & Mistral-Ours & 100 & 95\\
         \bottomrule
    \end{tabular}
    }
    \vspace{-3mm}
    \label{tab:task_success}
\end{table}

The vision module receives RGB-D input from the HSR head camera, performs object recognition using pretrained YOLO \cite{yolo}, and estimates object poses. 
For feasibility verification, we use the reachability graph from \cite{ly2023r}, a sampling-based approach that efficiently checks for paths from a starting point to a goal while ensuring feasible configurations for the robot reaching towards the target object. The graph sampled 200 nodes in our designated workspace, with whole-body motion planning. The timeout is set to 4 cycles searching with graph enhancement.

To verify applicability, the LLM model for our approach is fine-tuned from Mistral 7B \cite{jiang2023mistral} and Deepseek 8B \cite{guo2025deepseek}, state-of-the-art lightweight LLM models, which are suitable for onboard deployment. The fine-tuning data includes 300 samples covering all failure categories outlined in Table \ref{tab:failedcases}. Before execution, the robot is provided with a map of the room, e.g., large furniture like tables or drawers, and a list of motion APIs shared between the planner and the controller.

 To assess the performance of our method, we use three metrics. 1) \textit{Task planning success rate}: the ratio of the no-failure trials where the robot successfully generates the action sequence by understanding the assigned task. 2) \textit{Failure explanation success rate}: the ratio of the trials where the robot successfully generates the reason for not being able to find the action sequence. 3) \textit{Failure recovery success rate}: the ratio of the trials where the robot successfully generates the recovery plan in response to human guidance. 

We compare \textit{Mistral-Ours} and \textit{Deepseek-Ours} with few-shot prompting baselines, \textit{Mistral-Prompt} and \textit{Deepseek-Prompt}. The evaluation samples follow the same format as the dataset for fine-tuning, as discussed in Section \ref{sec:data}.  

\subsubsection{Comparison of Different Methods in Feasible Tasks}

\begin{figure}[ht]
\vspace{1mm}
    \centering
    \includegraphics[width=0.5\textwidth]{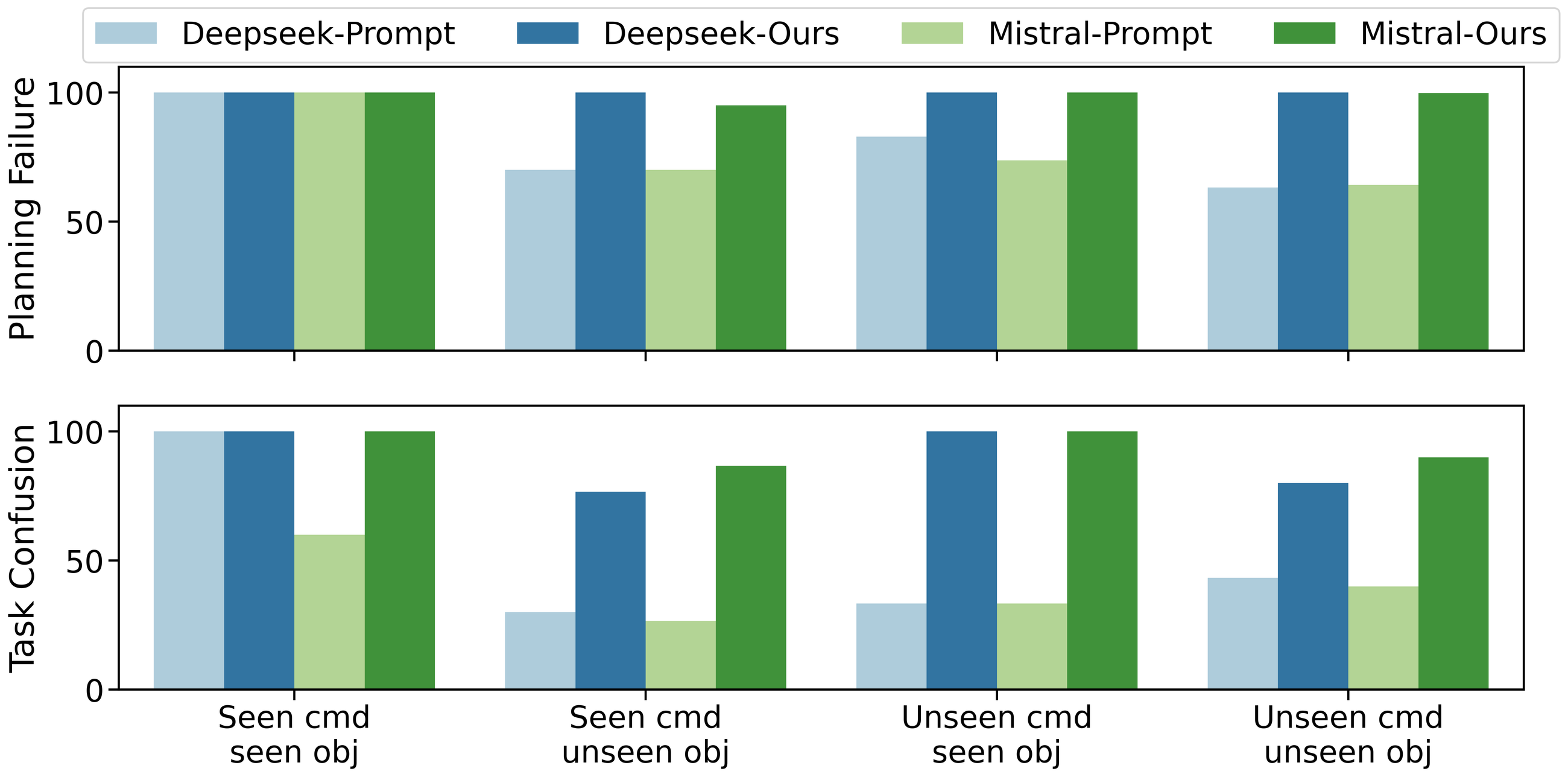}

    \vspace{-1mm}
    \caption{Success rates of failure explanation. }
    \vspace{-5mm}
    \label{fig:failuredetection}
\end{figure}

In our first experiment, we assume successful outcomes in both object detection and robotic motion feasibility by placing the requested object in the camera view and within the reach of the robot. In this set, the planner is solely required to generate the appropriate action sequences based on human inputs. An example of such conversations is provided in Fig. \ref{fig:examples}a.

Despite explicit instructions to produce solely an action sequence, \textit{Mistral-Prompt} and \textit{Deepseek-Prompt} often generate extraneous text.
For a fairer comparison, we only extracted the action sequence when determining the success rate, and ignored the lengthy explanation and slightly misused words (e.g., `pick up' vs `pick') for the action API interface. As presented in the \textit{w/o failure} column of Table \ref{tab:task_success}, the results demonstrate a superior performance of our models, achieving 100\% success in scenarios with seen commands and objects, and maintaining high effectiveness even with unseen objects or commands. In contrast, \textit{Mistral-Prompt} and  \textit{Deepseek-Prompt} experienced performance drops in scenarios involving unseen elements, with more sensitivity over unseen textual commands.  It is also observed that \textit{Deepseek-Prompt} behaves better than \textit{Mistral-Prompt}, while \textit{Deepseek-Ours} has slightly lower success rates than \textit{Mistral-Ours}.

\subsubsection{Evaluation of Robustness in Failure Explanation}
In this test, we evaluate the model’s ability to explicitly describe the cause of planning failures during the recovery process.
The failure experiments include all categories outlined in Table \ref{tab:failedcases}, with examples of `planning failure' detection in Fig. \ref{fig:examples}b,c. We only include `execution failure' in failure recovery experiments (Section \ref{ssec:exp_failurerecovery}), as it is a failure description that is triggered from the controller level.

Figure \ref{fig:failuredetection} shows that our method achieved near-perfect success rates in scenarios with seen commands and seen/unseen objects, underscoring its reliability in familiar settings. It is observed that `Task confusion' failure detection obtains lower success rates than multimodal failure detection. This is mainly due to the `ambiguous task' experiments, where the robot fails to distinguish object categories (e.g., fruits, drinks) from the list of detected objects. 
Our method increases the success rates of failure explanation by more than twice in these cases, proving that fine-tuning triggers the domain-specific understanding of the robots for which factor (e.g., user preference) to consider during planning.

\subsubsection{Evaluation of Robustness in Failure Recovery}
\label{ssec:exp_failurerecovery}
We further evaluate the framework’s ability to replan based on human instructions. In each trial, a detected failure is inserted into \textit{\textless history\textgreater}, and the models are expected to generate the appropriate recovery action.

\begin{figure}[t]
\vspace{1mm}
    \centering
    \includegraphics[width=0.5\textwidth]{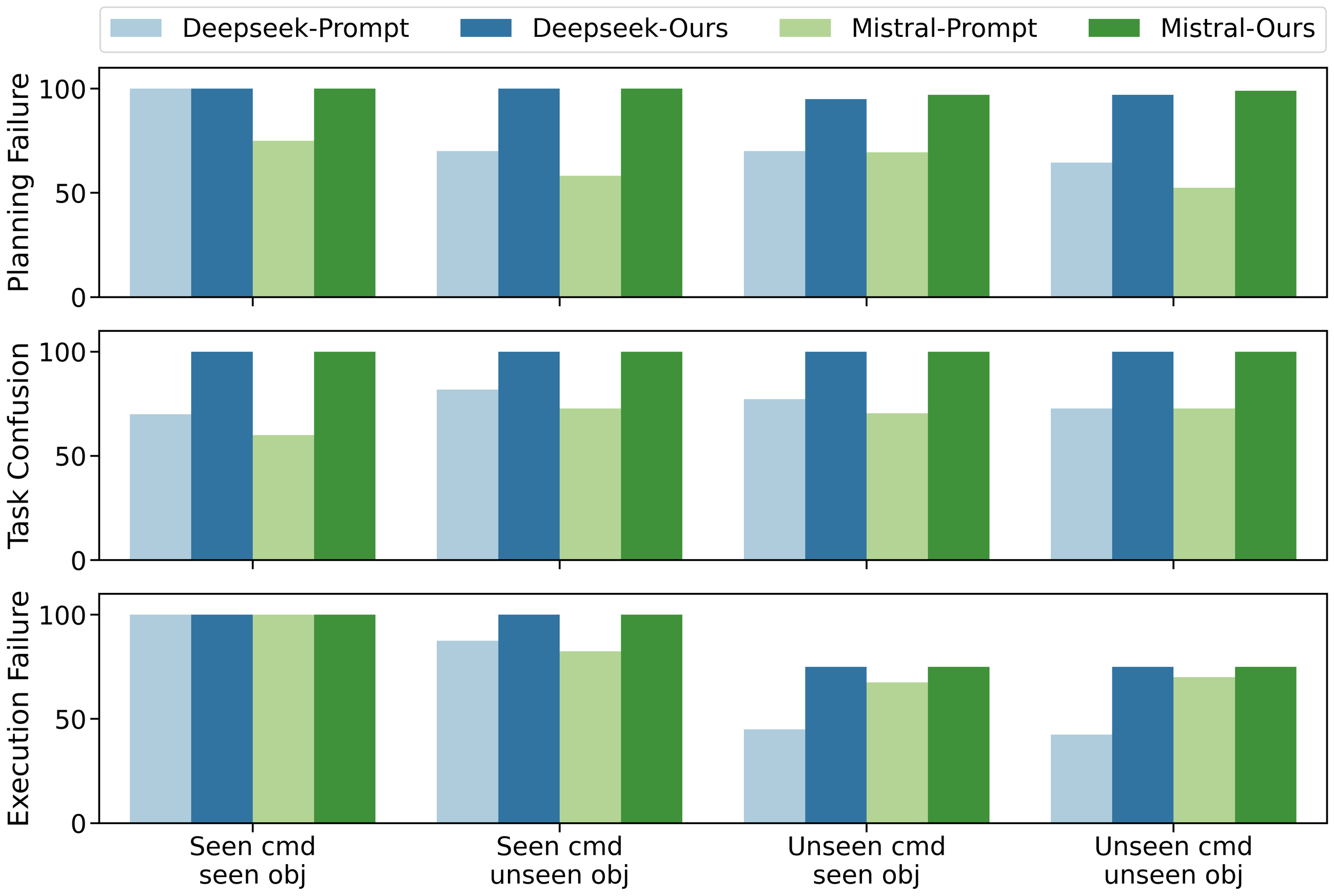}
    \vspace{-5mm}
    
    \caption{Success rates of failure recovery. }
    \label{fig:failurerecovery}
    \vspace{-5mm}
\end{figure}

Fig. \ref{fig:failurerecovery} demonstrates our framework's performance in replanning in various scenarios. While `task confusion' mainly processes user preference (e.g., user prefers the robot to get a Coke to a 7-Up), `planning failure' and `execution failure' cases assess the ability of mapping language to the list of pre-defined motions (e.g., output `open cupboard' in response to `the object is in the cupboard'). The lower success rates in recovering from `execution failure' are due to the longer sequence of recovery than what was provided in the fine-tuning data. In general, prompting methods fails to embody LLM with sufficient robot-domain understanding. Our method significantly increases the performance of LLM models in robotic task failure recovery, indicating robustness in real-world conditions where adaptability and responsiveness to failures are crucial for operational success.

The \textit{w/ failure} column of Table \ref{tab:task_success} reports success rates in scenarios where the planner must succeed in the two-step process of failure detection followed by failure recovery. The results reveal a significantly increased gap between prompt-based models and our approach as opposed to \textit{w/o failure} cases, underscoring the effectiveness of our method in handling multi-step interaction and ensuring task completion.

\subsection{Evaluation of Scalability Across Tasks}
\label{sec:scalability}
We evaluate the scalability of our approach using 101 task instructions from SayCan \cite{ahn2022can}. To accommodate the task variants, we fine-tuned Mistral with the tasks outlined in Table \ref{tab:testcase}, which covers the expected actions required for the dataset solutions.
We also introduce the \textit{Execution success rate} metric, defined as the ratio of the trials where the robot successfully executes the planned actions. 
Since SayCan \cite{ahn2022can} does not support failure recovery, we report only results from cases without recovery. 
Generative results from \textit{PaLM 8B} \cite{chowdhery2022palmscalinglanguagemodeling} are also reported as a lightweight LLM baseline. 

\begin{table}[h]
    \centering
    \caption{Tasks with Corresponding Expected Outcome.}
\vspace{-1mm}
    
    \begin{tabular}{p{2cm}|p{5.7cm}}
        \toprule
            Case & Expected plan \\
         
         \midrule
         Pick object & Pick object\\
         Go to destination & Go to destination\\
         Fetch me & Pick object, Go back, Place object\\
         Put away & Pick object, Go to destination, Place object\\
         Put in drawer & Open drawer, Pick object, Place in drawer, Close drawer \\
         
         \bottomrule
         
    \end{tabular}
    \label{tab:testcase}
\vspace{-2mm}
\end{table}

Table \ref{tab:saycantest} presents the tested results of the frameworks with the dataset from \textit{Saycan}. As a 540B model, the \textit{SayCan} result explains its excellent decision-making capability with throughout understanding of the world. Its trained affordance value provides sufficient embodied knowledge, ensuring a high success rate in execution. 
Besides, we note that
prompting with \textit{PaLM-8B} only successfully plans 38\% cases. \textit{Mistral-Prompt} with our modular structure helps the LLM model to gain embodied intelligence, with the planning success rate to 59\%.  It is observed that \textit{Mistral-Prompt}'s failure cases come from its lack of sense of the operating environment. For example, some of the results tell the robot to `go to store' to pick something up, despite being told that it is working in a domestic environment. It is proven that with 7B parameters, our \textit{InteLiPlan} obtains similar results in comparison to the state-of-the-art 504B SayCan model with an 83\% planning success rate.

The execution success rate remains relative to the planning success rate in \textit{Mistral-Prompt} and \textit{Mistral-Ours}, given the same vision and reachability modules check for the system. Having the reachability graph checking whole-body feasibility explicitly instead of using skill probabilities like SayCan, our approach also increases the chances of successfully executing the plan once generated. This is shown by the reduced difference in plan and execute success rates between \textit{SayCan} and our pipeline. On the other hand, as we use a sampling-based approach, there is a risk of not finding a path even if there is one exists, leading to failures from the planning level. This is where the failure recovery comes in. Another solution is to use more samples in the graph such that they sufficiently provide the reachability knowledge. Since \textit{InteLiPlan} is a plug-and-play system, it can seamlessly integrate future state-of-the-art feasibility modules.

\begin{table}[h]
    \caption{Plan and Execution Success Rates (\%) of the methods for 101 SayCan task descriptions.}
\vspace{-1mm}
    
    \centering
    \begin{tabular}{c|c|c}
        \toprule
         Methods & Plan  & Execute \\
         \midrule
         PaLM-8B & 38 & n/a \\ 
        PaLM-504B SayCan & 84 & 74 \\
         Mistral-7B Prompt& 59 & 58\\
        Mistral-7B Ours& 83  & 82 \\
         \bottomrule
    \end{tabular}
\vspace{-2mm}
    \label{tab:saycantest}
\end{table}

\subsection{Evaluation of Cross-Embodiedment with Real Robots}

We validate the cross-embodiment feature of the system on two different physical robot platforms that are different in both camera placements and the number of degrees of freedom (DoF). The skill sets are named identically in the two platforms to allow direct transfer without retuning from the pre-trained model's output to action execution. In both robot setups, InteLiPlan is processed entirely onboard with a single NVIDIA GeForce RTX 2070, verifying onboard computing capability. We tested a pre-trained model on these platforms:
\begin{itemize}
    \item \textit{Toyota Human Support Robot (HSR)}: 11-DoF robot. The vision module takes the signal from the Asus Xtion PRO Live stereo camera on the head of the robot.
    \item \textit{Anymal D-Unitree Z1 (Anytree)}: 12-DoF robot (where the Anymal D is simplified with a planar base). The vision module takes the signal from the RealSense D435 on the wrist of the end effector.
\end{itemize}

We implemented 3 sets of experiments, with 10 trials each:
\begin{itemize}
    \item \textit{No-failure task}: the robot is asked to complete a task by processing a high-level command, and the objects are guaranteed to be within the observation and reachable.
    \item \textit{1-step failure task}: we implement the planning failure from vision, where the targeted object is not within its observation, the robot is expected to report the failure to the user and replan upon guidance.
    \item \textit{Multi-step failure task}: the robot must detect failures and replan twice to be able to complete the task. Specifically, we implement the failure from vision first, then after the targeted object is found, it is out of reach and the robot should ask again to complete the task.
\end{itemize}

\begin{figure}[t]
\vspace{1mm}
    \centering
    \begin{subfigure}[b]{0.492\textwidth}
        \centering
        \includegraphics[width=\textwidth]{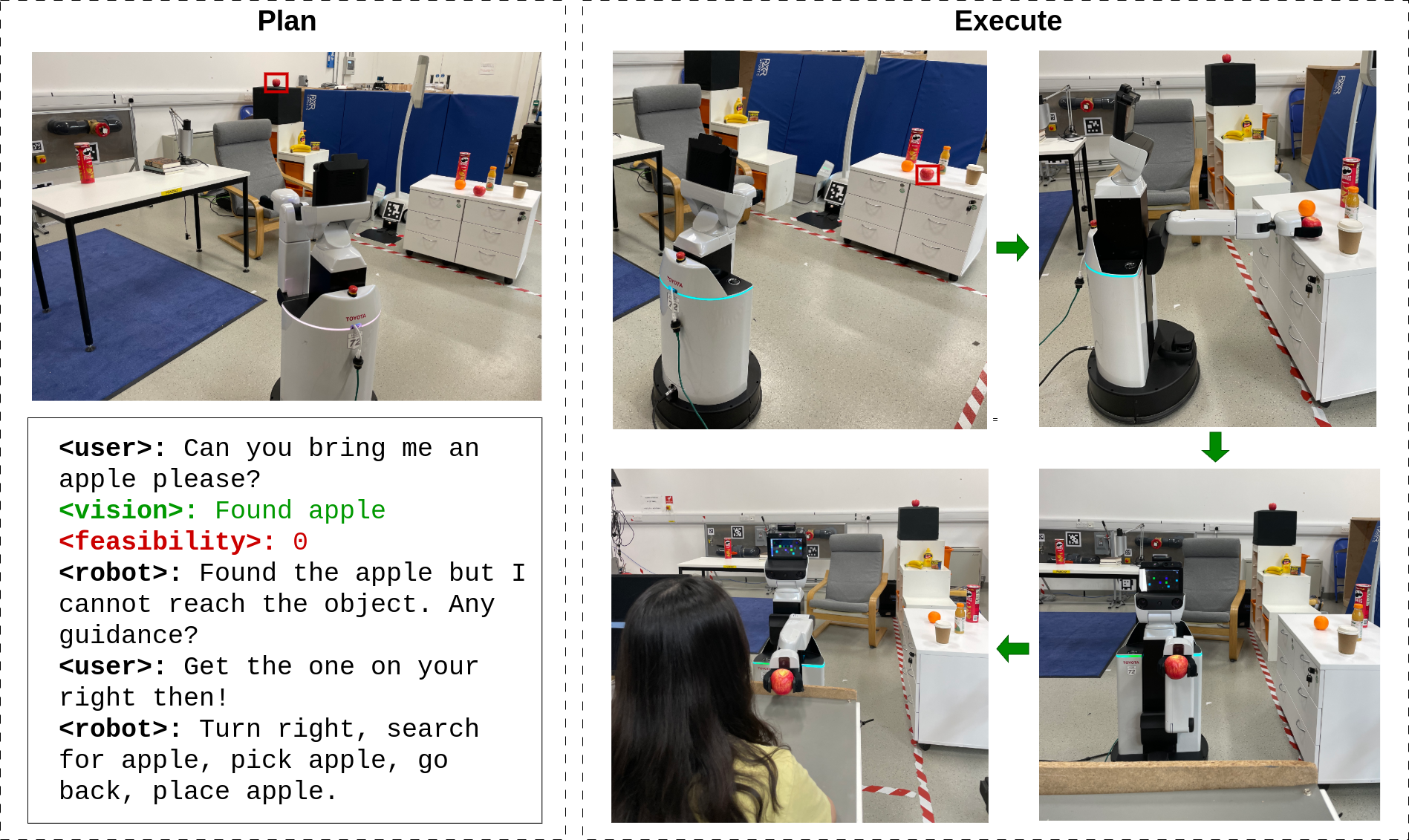}
        \caption{Failure from feasibility}
    \end{subfigure}
    \centering
    \par \medskip
    \begin{subfigure}[b]{0.492\textwidth}
        \centering
        \includegraphics[width=\textwidth]{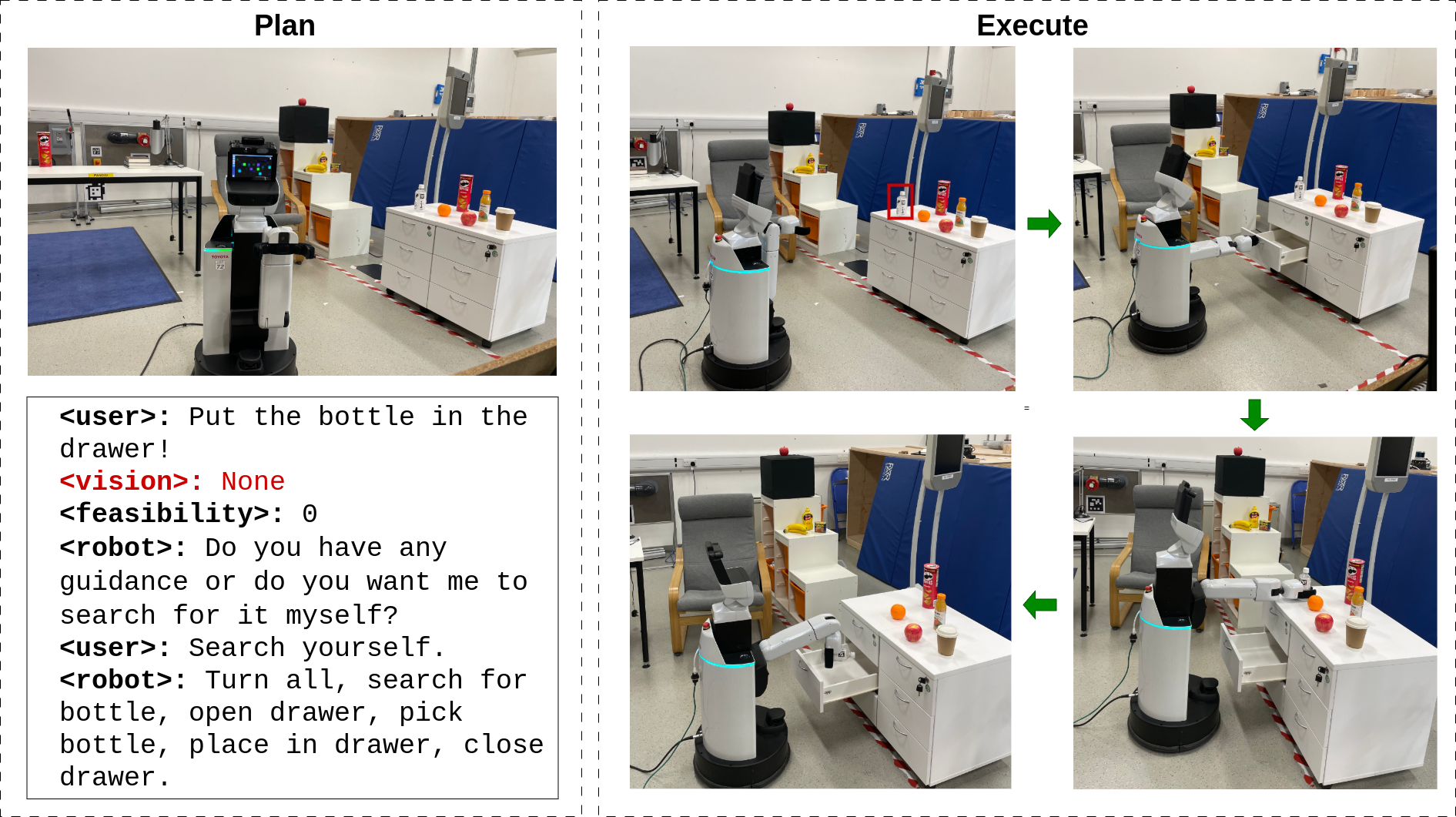}
        \caption{Failure from vision}
    \end{subfigure}
    
    \caption{Demonstration of the interactive failure recovery capability on the physical HSR. (a) captures HITL replanning with success in vision but failure in action feasibility, while (b) recovers from being unable to find the targeted object.}
    \label{fig:physicalhsr_fail}
    \vspace{-4mm}
\end{figure}

\begin{table}[ht]
\vspace{-1mm}
    \caption{Success Rates (\%) of Task Planning.}
\vspace{-1mm}
    \centering
    \begin{tabular}{c|c|c|c}
        \toprule
        Robot & w/o failure  & 1-step failure & Multi-step failure \\
         \midrule

        HSR & 100 & 100 & 90 \\ 
        Anytree & 100 & 100 & 90\\
         \bottomrule
    \end{tabular}
    
    \label{tab:physical_success}
\vspace{-1mm}
    
\end{table}
Table \ref{tab:physical_success} recorded the success rates of InteLiPlan in generating the correct sequences of action upon the corresponding case. The success rates were also affected by the success of vision and feasibility modules, leading to the failed cases. The result demonstrates that, with one model, the planned
action sequences generalize effectively across different embodiments without additional fine-tuning.
Fig. \ref{fig:physicalhsr} and Fig. \ref{fig:physicalhsr_fail} showcase \textit{InteLiPlan} results in real-world settings, which include tasks with and without failure recovery. The multi-step failure verifies that our method enables autonomy with seamless end-user intervention. 

Table \ref{tab:time} records the averaged planning time of the system over the trials, which is divided into vision, feasibility and planner queries processing time. In total, the model can plan an executable action sequence in less than 7s with onboard resources. The feasibility check module, which takes up the longest processing time, can be relaxed with fewer nodes in the reachability graph. Notably, using YOLO instead of VLM for vision input significantly contributes to real-time and onboard processing capability of the pipeline.

\begin{table}[ht]
\vspace{-1mm}
    \caption{Breakdown of average onboard planning time(s).}
\vspace{-1mm}
    \centering
    \begin{tabular}{c|c|c|c}
        \toprule
         Robot & Vision query & Feasibility query  & Planner query \\
         \midrule
         HSR & 6e-6 $\pm$ 2e-7 & 5 $\pm$ 0.14 & 1.5 $\pm$ 0.07 \\
         Anytree & 8e-6 $\pm$ 2e-7 & 5.5 $\pm$ 0.12 & 1.3 $\pm$ 0.08\\
         \bottomrule
    \end{tabular}
    \label{tab:time}
\vspace{-1mm}
\end{table}
The experiment validates the cross-embodiedment feature that requires no extra finetuning, further strengthening that InteLiPlan's planning capability is universal to different robots by not using robot-specific data. As opposed to vision-language-action models \cite{ma2024survey}, the approach allows direct transfer to robot with different camera views.

\section{Conclusion}
\label{sec:conclusion}

We presented \textit{InteLiPlan}, an interactive lightweight LLM-based robotics planner for reliable and robust autonomy in domestic environments. Our framework employs a conversation-like format between internal modules and facilitates human-robot interaction, allowing the system to reason about failures through a multimodal input formulation. By incorporating a human in the loop, with human-like text understanding capabilities of LLMs, the robot gains the ability to replan based on instructions, effectively utilizing human input for dynamic adjustments. We performed extensive evaluations to investigate common failures that happen in real scenarios, where human instruction is required. The interactive behaviors guarantee interpretability and reliability for robots operating in domestic environments.

This paper reduces the effort of defining traditional task planning domains in robotics, provides an intuitive, robot-independent data structure, and improves applicability in resource-constrained settings.
Notably, deploying our approach with the lightweight Mistral 7B model achieves both comparable results with the SOTA baseline and --notably to our approach-- real-time onboard computing. 
InteLiPlan approaches the performance of the 504B PaLM-SayCan system while operating fully onboard with a lightweight 7B model. Our system was validated on the physical HSR and Anymal D-Unitree Z1 robots, verifying the cross-embodiedment capability across different DoFs and camera views.

\textit{Limitations and Future Work} - 
Our framework is a multimodal system with a feasibility check, but it currently lacks
motion-level replanning feature.
Integrating low-level reactive behaviors with high-level replanning in a
dual-process manner could improve failure recovery in real-world scenarios.
Furthermore, the sampled path from the reachability graph could be extended into a motion planning module,
enabling a full task and motion planning stack.

\clearpage

\bibliographystyle{IEEEtran}
\bibliography{bibliography}

@IEEEtranBSTCTL{IEEEexample:BSTcontrol,
  CTLuse_article_number     = "yes",
  CTLuse_paper              = "yes",
  CTLuse_forced_etal        = "yes",
  CTLmax_names_forced_etal  = "10",
  CTLnames_show_etal        = "10",
}

@misc{kim2025finetuningvisionlanguageactionmodelsoptimizing,
      title={Fine-Tuning Vision-Language-Action Models: Optimizing Speed and Success}, 
      author={Moo Jin Kim and Chelsea Finn and Percy Liang},
      year={2025},
      eprint={2502.19645},
      archivePrefix={arXiv},
      primaryClass={cs.RO},
      url={https://arxiv.org/abs/2502.19645}, 
}

@misc{intelligence2025pi05visionlanguageactionmodelopenworld,
      title={$\pi_{0.5}$: a Vision-Language-Action Model with Open-World Generalization}, 
      author={Physical Intelligence and Kevin Black and Noah Brown and James Darpinian and Karan Dhabalia and Danny Driess and Adnan Esmail and Michael Equi and Chelsea Finn and Niccolo Fusai and Manuel Y. Galliker and Dibya Ghosh and Lachy Groom and Karol Hausman and Brian Ichter and Szymon Jakubczak and Tim Jones and Liyiming Ke and Devin LeBlanc and Sergey Levine and Adrian Li-Bell and Mohith Mothukuri and Suraj Nair and Karl Pertsch and Allen Z. Ren and Lucy Xiaoyang Shi and Laura Smith and Jost Tobias Springenberg and Kyle Stachowicz and James Tanner and Quan Vuong and Homer Walke and Anna Walling and Haohuan Wang and Lili Yu and Ury Zhilinsky},
      year={2025},
      eprint={2504.16054},
      archivePrefix={arXiv},
      primaryClass={cs.LG},
      url={https://arxiv.org/abs/2504.16054}, 
}

@misc{kim2024openvlaopensourcevisionlanguageactionmodel,
      title={OpenVLA: An Open-Source Vision-Language-Action Model}, 
      author={Moo Jin Kim and Karl Pertsch and Siddharth Karamcheti and Ted Xiao and Ashwin Balakrishna and Suraj Nair and Rafael Rafailov and Ethan Foster and Grace Lam and Pannag Sanketi and Quan Vuong and Thomas Kollar and Benjamin Burchfiel and Russ Tedrake and Dorsa Sadigh and Sergey Levine and Percy Liang and Chelsea Finn},
      year={2024},
      eprint={2406.09246},
      archivePrefix={arXiv},
      primaryClass={cs.RO},
      url={https://arxiv.org/abs/2406.09246}, 
}

@article{chowdhery2022palmscalinglanguagemodeling,
  title={Palm: Scaling language modeling with pathways},
  author={Chowdhery, Aakanksha and Narang, Sharan and Devlin, Jacob and Bosma, Maarten and Mishra, Gaurav and Roberts, Adam and Barham, Paul and Chung, Hyung Won and Sutton, Charles and Gehrmann, Sebastian and others},
  journal={Journal of Machine Learning Research},
  volume={24},
  number={240},
  pages={1--113},
  year={2023}
}

@article{brohan2022rt,
  title={Rt-1: Robotics transformer for real-world control at scale},
  author={Brohan, Anthony and Brown, Noah and Carbajal, Justice and Chebotar, Yevgen and Dabis, Joseph and Finn, Chelsea and Gopalakrishnan, Keerthana and Hausman, Karol and Herzog, Alex and Hsu, Jasmine and others},
  journal={arXiv preprint arXiv:2212.06817},
  year={2022}
}

@inproceedings{brohan2023rt,
  title={Rt-2: Vision-language-action models transfer web knowledge to robotic control},
  author={Zitkovich, Brianna and Yu, Tianhe and Xu, Sichun and Xu, Peng and Xiao, Ted and Xia, Fei and Wu, Jialin and Wohlhart, Paul and Welker, Stefan and Wahid, Ayzaan and others},
  booktitle={Conference on Robot Learning},
  pages={2165--2183},
  year={2023},
  organization={PMLR}
}

@inproceedings{silver2022pddl,
  title={{PDDL Planning with Pretrained Large Language Models}},
  author={Silver, Tom and Hariprasad, Varun and Shuttleworth, Reece S and Kumar, Nishanth and Lozano-P{\'e}rez, Tom{\'a}s and Kaelbling, Leslie Pack},
  booktitle={NeurIPS 2022 Foundation Models for Decision Making Workshop},
  year={2022}
}

@inproceedings{liang2023code,
  title={Code as policies: Language model programs for embodied control},
  author={Liang, Jacky and Huang, Wenlong and Xia, Fei and Xu, Peng and Hausman, Karol and Ichter, Brian and Florence, Pete and Zeng, Andy},
  booktitle={2023 IEEE International Conference on Robotics and Automation (ICRA)},
  pages={9493--9500},
  year={2023},
  organization={IEEE}
}

@article{kaddour2023challenges,
  title={Challenges and applications of large language models},
  author={Kaddour, Jean and Harris, Joshua and Mozes, Maximilian and Bradley, Herbie and Raileanu, Roberta and McHardy, Robert},
  journal={arXiv preprint arXiv:2307.10169},
  year={2023}
}

@inproceedings{ahn2022can,
  title={Do as i can, not as i say: Grounding language in robotic affordances},
  author={Brohan, Anthony and Chebotar, Yevgen and Finn, Chelsea and Hausman, Karol and Herzog, Alexander and Ho, Daniel and Ibarz, Julian and Irpan, Alex and Jang, Eric and Julian, Ryan and others},
  booktitle={Conference on robot learning},
  pages={287--318},
  year={2023},
  organization={PMLR}
}

@article{lin2023text2motion,
  title={Text2motion: From natural language instructions to feasible plans},
  author={Lin, Kevin and Agia, Christopher and Migimatsu, Toki and Pavone, Marco and Bohg, Jeannette},
  journal={Autonomous Robots},
  volume={47},
  number={8},
  pages={1345--1365},
  year={2023},
  publisher={Springer}
}

@inproceedings{pallagani2022plansformer,
  title={Plansformer: Generating Symbolic Plans using Transformers},
  author={Pallagani, Vishal and Muppasani, Bharath and Murugesan, Keerthiram and Rossi, Francesca and Horesh, Lior and Srivastava, Biplav and Fabiano, Francesco and Loreggia, Andrea},
  booktitle={NeurIPS 2023 Workshop on Generalization in Planning}
}

@article{lynch2023interactive,
  title={Interactive language: Talking to robots in real time},
  author={Lynch, Corey and Wahid, Ayzaan and Tompson, Jonathan and Ding, Tianli and Betker, James and Baruch, Robert and Armstrong, Travis and Florence, Pete},
  journal={IEEE Robotics and Automation Letters},
  year={2023},
  publisher={IEEE}
}

@INPROCEEDINGS{yolo,
  author={Redmon, Joseph and Divvala, Santosh and Girshick, Ross and Farhadi, Ali},
  booktitle={2016 IEEE Conference on Computer Vision and Pattern Recognition (CVPR)}, 
  title={You Only Look Once: Unified, Real-Time Object Detection}, 
  year={2016},
  volume={},
  number={},
  OPTpages={779-788},
  OPTdoi={10.1109/CVPR.2016.91}
}

@techreport{vemprala2023chatgpt,
author = {Vemprala, Sai and Bonatti, Rogerio and Bucker, Arthur and Kapoor, Ashish},
title = {ChatGPT for Robotics: Design Principles and Model Abilities},
institution = {Microsoft},
year = {2023},
month = {February},
OPTurl = {https://www.microsoft.com/en-us/research/publication/chatgpt-for-robotics-design-principles-and-model-abilities/},
number = {MSR-TR-2023-8},
}

@inproceedings{ly2023r,
  title={R-lgp: A reachability-guided logic-geometric programming framework for optimal task and motion planning on mobile manipulators},
  author={Ly, Kim Tien and Semenov, Valeriy and Risiglione, Mattia and Merkt, Wolfgang and Havoutis, Ioannis},
  booktitle={2024 IEEE International Conference on Robotics and Automation (ICRA)},
  pages={14917--14923},
  year={2024},
  organization={IEEE}
}

@inproceedings{yang2022sequence,
  title={Sequence-Based Plan Feasibility Prediction for Efficient Task and Motion Planning},
  author={Yang, Zhutian and Garrett, Caelan and Lozano-Perez, Tomas and Kaelbling, Leslie and Fox, Dieter},
  booktitle={Robotics science and systems},
  year={2023}
}

@inproceedings{huang2022language,
  title={Language models as zero-shot planners: Extracting actionable knowledge for embodied agents},
  author={Huang, Wenlong and Abbeel, Pieter and Pathak, Deepak and Mordatch, Igor},
  booktitle={International Conference on Machine Learning},
  pages={9118--9147},
  year={2022},
  organization={PMLR}
}

@inproceedings{shirasaka2023self,
  title={Self-recovery prompting: Promptable general purpose service robot system with foundation models and self-recovery},
  author={Shirasaka, Mimo and Matsushima, Tatsuya and Tsunashima, Soshi and Ikeda, Yuya and Horo, Aoi and Ikoma, So and Tsuji, Chikaha and Wada, Hikaru and Omija, Tsunekazu and Komukai, Dai and others},
  booktitle={2024 IEEE International Conference on Robotics and Automation (ICRA)},
  pages={17395--17402},
  year={2024},
  organization={IEEE}
}

@inproceedings{huang2022inner,
  title={Inner Monologue: Embodied Reasoning through Planning with Language Models},
  author={Huang, Wenlong and Xia, Fei and Xiao, Ted and Chan, Harris and Liang, Jacky and Florence, Pete and Zeng, Andy and Tompson, Jonathan and Mordatch, Igor and Chebotar, Yevgen and others},
  booktitle={Conference on Robot Learning},
  pages={1769--1782},
  year={2023},
  organization={PMLR}
}

@article{ahn2024vader,
  title={VADER: Visual Affordance Detection and Error Recovery for Multi Robot Human Collaboration},
  author={Ahn, Michael and Arenas, Montserrat Gonzalez and Bennice, Matthew and Brown, Noah and Chan, Christine and David, Byron and Francis, Anthony and Gonzalez, Gavin and Hessmer, Rainer and Jackson, Tomas and others},
  journal={arXiv preprint arXiv:2405.16021},
  year={2024}
}

@inproceedings{joublin2024copal,
  title={CoPAL: corrective planning of robot actions with large language models},
  author={Joublin, Frank and Ceravola, Antonello and Smirnov, Pavel and Ocker, Felix and Deigmoeller, Joerg and Belardinelli, Anna and Wang, Chao and Hasler, Stephan and Tanneberg, Daniel and Gienger, Michael},
  booktitle={2024 IEEE International Conference on Robotics and Automation (ICRA)},
  pages={8664--8670},
  year={2024},
  organization={IEEE}
}

@inproceedings{wang2023describe,
  title={Describe, explain, plan and select: interactive planning with large language models enables open-world multi-task agents},
  author={Wang, Zihao and Cai, Shaofei and Chen, Guanzhou and Liu, Anji and Ma, Xiaojian and Liang, Yitao and CraftJarvis, Team},
  booktitle={Proceedings of the 37th International Conference on Neural Information Processing Systems},
  pages={34153--34189},
  year={2023}
}

@article{jiang2023mistral,
  title={Mistral 7B},
  author={Jiang, Albert Q and Sablayrolles, Alexandre and Mensch, Arthur and Bamford, Chris and Chaplot, Devendra Singh and Casas, Diego de las and Bressand, Florian and Lengyel, Gianna and Lample, Guillaume and Saulnier, Lucile and others},
  journal={arXiv preprint arXiv:2310.06825},
  year={2023}
}

@article{brown2020language,
  title={Language models are few-shot learners},
  author={Brown, Tom and Mann, Benjamin and Ryder, Nick and Subbiah, Melanie and Kaplan, Jared D and Dhariwal, Prafulla and Neelakantan, Arvind and Shyam, Pranav and Sastry, Girish and Askell, Amanda and others},
  journal={Advances in neural information processing systems},
  volume={33},
  pages={1877--1901},
  year={2020}
}

@article{wang2024large,
  title={Large language models for robotics: Opportunities, challenges, and perspectives},
  author={Wang, Jiaqi and Wu, Zihao and Li, Yiwei and Jiang, Hanqi and Shu, Peng and Shi, Enze and Hu, Huawen and Ma, Chong and Liu, Yiheng and Wang, Xuhui and others},
  journal={arXiv preprint arXiv:2401.04334},
  year={2024}
}

@article{zhang2023large,
  title={Large language models for human-robot interaction: A review},
  author={Zhang, Ceng and Chen, Junxin and Li, Jiatong and Peng, Yanhong and Mao, Zebing},
  journal={Biomimetic Intelligence and Robotics},
  pages={100131},
  year={2023},
  publisher={Elsevier}
}

@inproceedings{toussaint2015logic,
  title={Logic-geometric programming: an optimization-based approach to combined task and motion planning},
  author={Toussaint, Marc},
  booktitle={Proceedings of the 24th International Conference on Artificial Intelligence},
  OPTpages={1930--1936},
  year={2015}
}

@article{ren2023robots,
  title={Robots That Ask For Help: Uncertainty Alignment for Large Language Model Planners},
  author={Ren, Allen Z and Dixit, Anushri and Bodrova, Alexandra and Singh, Sumeet and Tu, Stephen and Brown, Noah and Xu, Peng and Takayama, Leila and Xia, Fei and Varley, Jake and others},
  journal={Proceedings of Machine Learning Research},
  volume={229},
  year={2023}
}

@inproceedings{raman2024cape,
  title={Cape: Corrective actions from precondition errors using large language models},
  author={Raman, Shreyas Sundara and Cohen, Vanya and Idrees, Ifrah and Rosen, Eric and Mooney, Raymond and Tellex, Stefanie and Paulius, David},
  booktitle={2024 IEEE International Conference on Robotics and Automation (ICRA)},
  pages={14070--14077},
  year={2024},
  organization={IEEE}
}

@inproceedings{liu2023reflect,
  title={REFLECT: Summarizing Robot Experiences for Failure Explanation and Correction},
  author={Liu, Zeyi and Bahety, Arpit and Song, Shuran},
  booktitle={Conference on Robot Learning},
  pages={3468--3484},
  year={2023},
  organization={PMLR}
}

@article{guo2025deepseek,
  title={Deepseek-r1: Incentivizing reasoning capability in llms via reinforcement learning},
  author={Guo, Daya and Yang, Dejian and Zhang, Haowei and Song, Junxiao and Zhang, Ruoyu and Xu, Runxin and Zhu, Qihao and Ma, Shirong and Wang, Peiyi and Bi, Xiao and others},
  journal={arXiv preprint arXiv:2501.12948},
  year={2025}
}

@article{hu2022lora,
  title={Lora: Low-rank adaptation of large language models.},
  author={Hu, Edward J and Shen, Yelong and Wallis, Phillip and Allen-Zhu, Zeyuan and Li, Yuanzhi and Wang, Shean and Wang, Lu and Chen, Weizhu and others},
  journal={International Conference on Learning Representations.},
  volume={1},
  number={2},
  pages={3},
  year={2022}
}

@article{weyssow2023exploring,
  title={Exploring parameter-efficient fine-tuning techniques for code generation with large language models},
  author={Weyssow, Martin and Zhou, Xin and Kim, Kisub and Lo, David and Sahraoui, Houari},
  journal={ACM Transactions on Software Engineering and Methodology},
  year={2023},
  publisher={ACM New York, NY}
}

@article{ding2023parameter,
  title={Parameter-efficient fine-tuning of large-scale pre-trained language models},
  author={Ding, Ning and Qin, Yujia and Yang, Guang and Wei, Fuchao and Yang, Zonghan and Su, Yusheng and Hu, Shengding and Chen, Yulin and Chan, Chi-Min and Chen, Weize and others},
  journal={Nature Machine Intelligence},
  volume={5},
  number={3},
  pages={220--235},
  year={2023},
  publisher={Nature Publishing Group UK London}
}

@article{mei2024replanvlm,
  title={Replanvlm: Replanning robotic tasks with visual language models},
  author={Mei, Aoran and Zhu, Guo-Niu and Zhang, Huaxiang and Gan, Zhongxue},
  journal={IEEE Robotics and Automation Letters},
  year={2024},
  publisher={IEEE}
}

@article{ma2024survey,
  title={A survey on vision-language-action models for embodied ai},
  author={Ma, Yueen and Song, Zixing and Zhuang, Yuzheng and Hao, Jianye and King, Irwin},
  journal={arXiv preprint arXiv:2405.14093},
  year={2024}
}
\end{document}